\documentclass{bmvc2k}
\usepackage{graphicx}
\usepackage{booktabs}
\usepackage{tabularx}
\usepackage{amsmath,amssymb} 

\title{Deep Covariance Descriptors for Facial Expression Recognition}

\addauthor{Naima Otberdout}{naima.otberdout@um5s.net.ma}{1}
\addauthor{Anis Kacem}{anis.kacem@imt-lille-douai.fr}{2}
\addauthor{Mohamed Daoudi}{mohamed.daoudi@imt-lille-douai.fr}{2}
\addauthor{Lahoucine Ballihi}{lahoucine.ballihi@um5.ac.ma}{1}
\addauthor{Stefano Berretti}{stefano.berretti@unifi.it}{3}

\addinstitution{
LRIT - CNRST URAC 29
Rabat IT Center 
Mohammed V University in Rabat
Faculty of sciences, Rabat, Morocco
}
\addinstitution{
 IMT Lille-Douai, Univ. Lille,\\
 CNRS, UMR 9189 CRIStAL,\\
 Lille, France
}
\addinstitution{
 Media Integration and Communication Center (MICC),
 University of Florence,\\
 Florence, Italy
}
\runninghead{Deep Covariance Descriptors for Facial Expression Recognition}{}

\begin{document}

\maketitle

\begin{abstract}
In this paper, covariance matrices are exploited to encode the deep convolutional neural networks (DCNN) features for facial expression recognition. The space geometry of the covariance matrices is that of Symmetric Positive Definite (SPD) matrices. By performing the classification of the facial expressions using Gaussian kernel on SPD manifold, we show that the covariance descriptors computed on DCNN features are more efficient than the standard classification with fully connected layers and softmax. By implementing our approach using the VGG-face and ExpNet architectures with extensive experiments on the Oulu-CASIA and SFEW datasets, we show that the proposed approach achieves performance at the state of the art for facial expression recognition. 
\end{abstract}

\section{Introduction}
\label{sec:intro}
Automatic analysis of facial expressions has been attractive in computer vision research since long time due to its wide spectrum of potential applications that go from human computer interaction to medical and psychological investigations, to cite a few. 
Similarly to other applications, for many years facial expression analysis has been addressed by designing hand-crafted low-level descriptors, either geometric (\emph{e.g.}, distances between landmarks) or appearance based (\emph{e.g.}, LBP, SIFT, HOG, etc.), with the aim of extracting suitable representations of the face. 
Higher order relations, like the covariance descriptor, have been also computed on raw data or low-level descriptors. 
Standard machine learning tools, like SVMs, have then been used to classify expressions. 
Now, the approach to address this problem has changed quite radically with Deep Convolutional Neural Networks (DCNNs). The idea here is to make the network learn the best features from large collections of data during a training phase.
However, one drawback of DCNNs is that they do not take into account the spatial relationships within the face. To overcome this issue, we propose to exploit globally and locally the network features extracted in different regions of the face. 
This yields a set of DCNN features per region. The question is how to encode them in a compact and discriminative representation for a more efficient classification than the one achieved globally by classical softmax. 
In this paper, we propose to encode face DCNN features in a covariance matrix. These matrices have shown to outperform first-order features in many computer vision tasks~\cite{tuzel:2006,tuzel:2008}. 
We demonstrate the benefits of this representation in facial expression recognition from static images or collections of static peak frames (\emph{i.e.}, frames where the expression reaches its maximum). In doing this, we exploit the space geometry of the covariance matrices as points on the symmetric positive definite (SPD) manifold. Furthermore, we use a valid positive definite Gaussian RBF kernel on this manifold to train a SVM classifier for expression classification. Implementing our approach with different network architectures, \emph{i.e.}, VGG-face~\cite{parkhi2015deep} and ExpNet~\cite{ding2017facenet2expnet}, and by a thorough set of experiments, we found that the classification of these matrices outperforms the classical softmax. 

\begin{figure}[!th]
\centering
\includegraphics[width=0.7\linewidth]{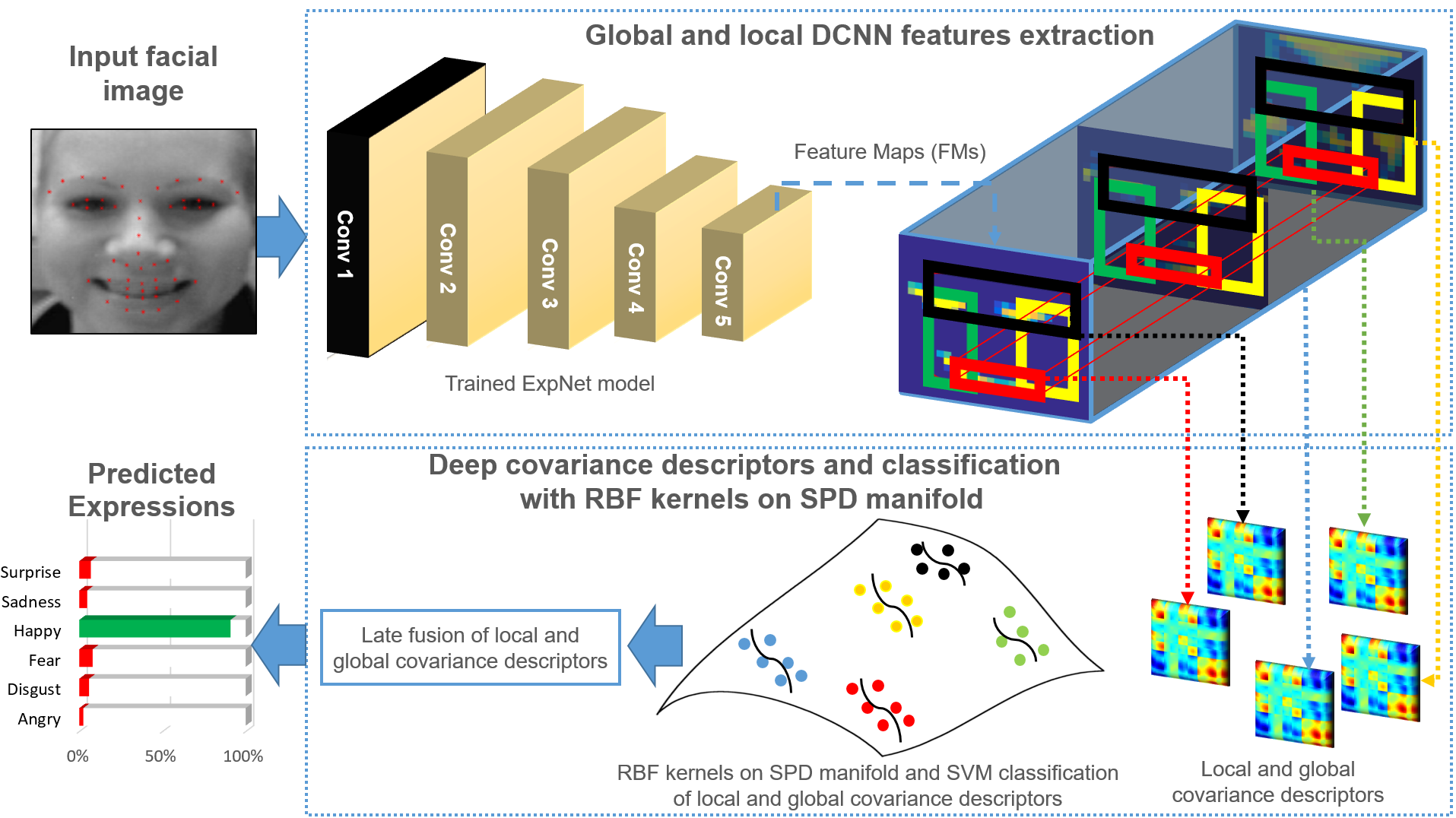}
\centering
\caption{Overview of the proposed method.}
\label{fig:method-overview}
\end{figure}

Overall, the proposed solution permits us to combine the geometric and appearance features enabling an effective description of  facial expressions at different spatial levels, while taking into account the spatial relationships within the face. An overview of the proposed solution is illustrated in Figure~\ref{fig:method-overview}. 
In summary, the main contributions of our work consist of: \textit{(i)} encoding DCNN features of the face by using covariance matrices; \textit{(ii)} encoding local DCNN features by local covariance descriptors; \textit{(iii)} classifying facial expressions using Gaussian kernel on the SPD manifold; \textit{(iv)} conducting an extensive experimental evaluation with two different architectures and comparing our results with state-of-the-art methods on two publicly available datasets.

The rest of the paper is organized as follows: In Section~\ref{sect:related-work}, we summarize the works that are most related to our solution, including facial expression recognition, and covariance descriptors; In Section~\ref{sect:approach}, we present our solution for facial feature extraction and, in Section~\ref{sect:dcnn-covariance}, we introduce the idea of DCNN covariance descriptors for expression classification. A comprehensive experimentation using the proposed approach on two publicly available benchmarks, and comparison with state-of-the-art solutions is reported in Section~\ref{sect:results}; Finally, conclusions and directions for future work are sketched in Section~\ref{sect:conclusion}.

\section{Related work}\label{sect:related-work}
The approach we propose in this paper is mainly related to the works on facial expression recognition and those on DCNNs combined with covariance descriptors. Accordingly, we first summarize relevant works using DCNN for facial expression, then we present some recent works that use covariance descriptors in conjunction with DCNN.

\textbf{DCNN for Facial Expression Recognition:} Motivated by the success of DCNN models in facial analysis tasks, several papers proposed to use them for both static and dynamic facial expression recognition~\cite{jung2015joint,Zafeiriou2017,mollahosseini2016going,ng2015deep}. 
However, the main reason behind the impressive performance of DCNNs is the availability of large-scale training datasets. As a matter of fact, in facial expression recognition, datasets are quite small, mainly for the difficulty of producing properly annotated images for training. 
To overcome such a problem, Ding et al.~\cite{ding2017facenet2expnet} proposed \textit{FaceNet2ExpNet}, where a regularization function helps to use the face information to train the facial expression classification net of static images. Facial expression recognition from still images using DCNN was also proposed in~\cite{mollahosseini2016going,ng2015deep,yu2015image}. All these methods use a similar strategy in the network architecture: multiple convolutional and pooling layers are used for feature extraction; fully connected ones, and softmax layers are used for classification. In~\cite{ofodile2017automatic}, the authors proposed a method for dynamic facial expression recognition that exploits deep features extracted at the last convolutional layer of a trained DCNN. They used a Gaussian Mixture Model (GMM) and Fisher vector encoding on the set of extracted features from videos to get a single vector representation of the video, which is fed into a SVM classifier to predict expressions. 

\textbf{DCNN and Covariance Descriptors:} Covariance features were first introduced by Tuzel et al.~\cite{tuzel:2006} for texture matching and classification. Bhattacharya et al.~\cite{bhattacharya:2016} constructed covariance matrices, which capture joint statistics of both low-level motion and appearance features extracted from a video. 
Dong et al.~\cite{dong2017deep} constructed a deep neural network, which embeds high dimensional SPD matrices into a more discriminative low dimensional SPD manifold. 
 In the context of face recognition from image sets, Wang et al.~\cite{wang2017discriminative} presented a Discriminative Covariance oriented Representation Learning (DCRL) framework to learn better image representations, which can closely match the subsequent image set modeling and classification. The framework constructs a feature learning network (\emph{e.g.}, a CNN) to project the face images into a target representation space. The network is trained with the goal of maximizing the discriminative ability of the set of covariance matrices computed in the target space. 
 In the dynamic facial expression recognition method proposed by Liu et al.~\cite{liu2014combining}, deep and hand-crafted features are extracted from each video clip to build three types of image set models, \emph{i.e.}, covariance matrix, linear subspace, and Gaussian distribution. Then, different Riemannian kernels are used, separately and combined, for classification.

To the best of our knowledge, compared to existing literature, our work is the first one that uses covariance descriptors in conjunction with DCNN for static expression recognition. 
\section{DCNN features}\label{sect:approach}
Given a set of $n_f$ face images $\mathcal{F} = \{f_{1}, f_{2},...f_{n_f} \}$ labeled with their corresponding expressions $\{y_{1}, y_{2},...y_{n_f}\}$, our goal is to find a high discriminative face representation allowing an efficient matching between faces and their expression labels. Motivated by the success of DCNNs in automatic extraction of non-linear features that are relevant to the problem at hand, we opt for this technique in order to encode the facial expression into Feature Maps (FMs). A covariance descriptor is then computed over these FMs and is considered for global face representation. We also extract four regions on the input face image around the eyes, mouth, and cheeks (left and right). 
By mapping these regions on the extracted deep FMs, we are able to extract local regions in these FMs that bring more accurate information about the facial expression. A local covariance descriptor is also computed for each local region. 

The first step to our approach is the extraction of non-linear features that encode well the facial expression in the input face image. In this work, we use two DCNN models, namely, \textit{VGG-face}~\cite{parkhi2015deep} and \textit{ExpNet}~\cite{ding2017facenet2expnet}.

\subsection{Global DCNN features}\label{sect:GDCNN}
VGG-Face is a DCNN model that is commonly used in facial analysis tasks. It consists of 16 layers trained on $2.6$M facial images of $2.6$K people for face recognition in the wild. This model has been also successfully used for expression recognition~\cite{ding2017facenet2expnet}. However, the model was trained for face identification, so it is expected to also encode information about the identity of the persons that should be filtered-out in order to capture person-independent facial expressions. This may deteriorate the discrimination of the expression model after fine-tuning, especially when it comes to dealing with small datasets, which is quite common in facial expression recognition.
To tackle this problem, Ding et al.~\cite{ding2017facenet2expnet} proposed ExpNet, which is a much smaller network containing only five convolutional layers and one fully connected layer. The training of this model is regularized by the VGG-face model.

Following Ding et al.~\cite{ding2017facenet2expnet}, we first fine-tune the VGG-face network on expression datasets by minimizing the cross entropy loss. This fine-tuned model is then used to regularize the ExpNet model. Because we are interested in facial feature extraction, we only consider the FMs at the last convolutional layer of the ExpNet model. In what follows, we will denote the set of extracted FMs from an input face image $f$ as $\Phi(f)=\{M_1, M_2, \cdots, M_{m}\}$,~where $\{M_i\}_{i=1}^{m}$ are the $m$ FMs at the last convolutional layer, and $\Phi(.)$ is the non-linear function induced by the employed DCNN architecture at this layer. 

\subsection{Local DCNN features}
In addition to using the global feature map $\Phi(f)$, we focus on specific regions extracted from this global feature map that are relevant for face expression analysis. 
To do so, we start by detecting a set of landmark points on the input facial image using the method proposed in~\cite{asthana2014incremental}. Four regions $\{R_{j}\}_{j=1}^4$ are then constructed around the eyes, mouth, and both cheeks using these points. By defining a pixel-wise mapping between the input face image and its corresponding FMs, we map the detected regions from the input face image to the global FMs. Indeed, a feature map $M_{i}$ is obtained by convolution of the input image with a linear filter, adding a bias term and then applying a non-linear function. Accordingly, units within a feature map will be connected to different regions $R_{j}$ on the input image. Based on this assumption, we can find a mapping between the coordinates of the input image and those of the output feature map. 
Specifically, for each point $p$ of coordinates $(x_{p}, y_{p})$ in the input face image $f$, we associate a feature $\phi_{p}(f, M_{i})$ in the feature map $M_{i}$ such that,
\begin{equation}
\label{eq:Mapping}
\phi_{p}(f, M_{i}) = M{_i}(\overline{s \times x_{p}}, \overline{s\times y_{p}}) \; ,
\end{equation}
\noindent
where $s$ is the map size ratio with respect to input size, and $\bar{(.)}$ is the rounding operation. It is worth noting that for both models used in this work, the input image and output maps have the same spatial extent. This is important to map landmarks position in the input image to the coordinates of convolutional feature maps. Using this pixel-wise mapping, we map each region $R_j$ formed by $r$ pixels $\{p_1,p_2,\cdots,p_r\}$ on the input image into the global FMs $\{M_i\}_{i=1}^{m}$ to obtain the corresponding local FMs $\Phi^{R_j}(f)= \{\phi_{p_1}(f, M_{i}),\phi_{p_2}(f, M_{i}),\cdots,\phi_{p_r}(f, M_{i})\}_{i=1}^{m}$.

\section{DCNN based covariance descriptors}\label{sect:dcnn-covariance}
Both our local and global non-linear features $\Phi(f)$ and $\{\Phi^{R_{j}}(f)\}_{j=1}^{4}$ can be directly used to classify the face images. However, motivated by the great success of covariance matrices in various recent works, we propose to compute covariance descriptors using these global and local features. In particular, a covariance descriptor is computed for each region $R_{j}$ across the corresponding local FMs $\Phi^{R_{j}}(f)$ yielding four covariance descriptors. A covariance descriptor is also computed on the global FMs $\Phi(f)$ extracted from the whole face $f$. 
In this way, we encode the correlation between the extracted non-linear features within different spatial levels, which results in an efficient, compact and more discriminative representation. Furthermore, covariance descriptors allow us to select local features and focus on local facial regions, which is not possible with fully connected and softmax layers. We can also note that the covariance descriptors are treated separately, then lately fused in the classifier. In what follows, we describe the processing for the global features $\Phi(f)$; the same steps hold for the covariance descriptors computed over the local features.

The extracted features $\Phi(f)$ are arranged in a $(m \times w \times h )$ tensor, where $w$ and $h$ denote the width and height of the feature maps, respectively, and $m$ is their number. Each feature map $M_i$ is vectorized into a $n$-dimensional vector with $n=w\times h$ to transform the input tensor to a set of $n$ observations stored in the matrix $[v_{1}, v_{2},...,v_{n}] \in \mathbb{R}^{m \times n}$. Each observation $\{v_i\}^n_{i=1} \in \mathbb{R}^{m}$ encodes the values of the pixel $i$ across all the $m$ feature maps. Finally, we compute the corresponding $(m\times m)$ covariance matrix, 
\begin{equation}
\label{eq:covariance}
C_{\Phi(f)} = \dfrac{1}{n-1}\sum^{n}_{i=1} (v_{i}-\mu)(v_{i}-\mu)^{T} \; ,
\end{equation}

\noindent
where $\mu$ is the mean of the feature vectors such that $\mu=\dfrac{1}{n}\sum^{n}_{i=1}v_{i}$. Covariance descriptors are mostly studied under a Riemannian structure of the space of symmetric positive definite matrices $Sym^{++}(m)$~\cite{tuzel:2006,jayasumana2015kernel,wang2017discriminative}.
Several metrics have been proposed to compare covariance matrices on $Sym^{++}(m)$, the most widely used is the Log-Euclidean Riemannian Metric (LERM)~\cite{arsigny2006log} since it has excellent theoretical properties with simple and fast computations. 
Formally, given two covariance descriptors $C_{\Phi(f_1)}$ and $C_{\Phi(f_2)}$ of two images $f_{1}$ and $f_{2}$, their log-Euclidean distance $d: (Sym^{++}(m) \times Sym^{++}(m)) \rightarrow \mathbb{R}^{+}$ is given by,
\begin{equation}
\label{eq:LERM}
d(C_{\Phi(f_1)},C_{\Phi(f_2)})=\| \log(C_{\Phi(f_1)})-\log(C_{\Phi(f_2)} ) \|_{\textrm{F}} \; ,
\end{equation}

\noindent
where $\|\cdot\|_{\textrm{F}}$ is the Frobenius norm, and $\log(.)$ is the matrix logarithm. 

\subsection{RBF Kernels for DCNN covariance descriptors classification} 
As discussed above, each face $f$ is represented by its global and local covariance descriptors that lie on the non-linear manifold $Sym^{++}(m)$.
The problem of recognizing expressions from facial images is then turned to classifying their covariance descriptors in $Sym^{++}(m)$. However, one should take into account the non-linearity of this space, where traditional machine learning techniques cannot be applied in a straightforward way. Accordingly, we exploit the the log-Euclidean distance mentioned in Eq.~\eqref{eq:LERM} between symmetric positive definite matrices to define the Gaussian RBF kernel $K : (Sym^{++}(m) \times Sym^{++}(m)) \rightarrow \mathbb{R}^{+}$,
\begin{equation}
\label{eq:kernel}
K(C_{\Phi(f_1)}, C_{\Phi(f_2)})=\exp (-\gamma d^2(C_{\Phi(f_1)}, C_{\Phi(f_2)})) \; ,
\end{equation}

\noindent
where $d(C_{\Phi(f_1)}, C_{\Phi(f_2)})$ is the log-Euclidean distance between $C_{\Phi(f_1)}$ and $C_{\Phi(f_2)}$. Conveniently for us, this kernel has been already proved to be a positive definite kernel for all $\gamma > 0$ \cite{jayasumana2015kernel}. 
This kernel is computed for the global covariance descriptor as well as for each local covariance descriptor yielding to five different kernels. Then, each kernel is fed, separately, to a SVM classifier that outputs a score per class. Finally, fusion is performed by multiplying or computing a weighted sum over the scores given by the different kernels. 
\section{Experimental results}\label{sect:results}
The effectiveness of the proposed approach in recognizing basic facial expressions has been evaluated in constrained and unconstrained (\emph{i.e.}, in-the-wild) settings using two publicly available datasets with different challenges:

\textbf{Oulu-CASIA dataset~\cite{zhao2011facial}:} Includes $480$ image sequences of $80$ subjects taken in a constrained environment with normal illumination conditions. For each subject, there are six sequences, one for each of the six basic emotion labels. Each sequence begins with a neutral facial expression and ends with the apex of the expression.
For both training and testing, we use the last three peak frames to represent the video resulting in $1440$ images. Following the same setting of the state-of-the-art, we conducted a ten-fold cross validation experiment, with subject independent splitting. 

\textbf{Static Facial Expression in the Wild (SFEW) dataset~\cite{dhall2015video}:} Consists of $1,322$ static images labeled with seven facial expressions (the six basic plus the neutral one). This dataset has been collected from real movies and targets spontaneous expression recognition in challenging, \emph{i.e.}, in-the-wild, environments. It is divided into training (891 samples), validation ($431$ samples), and test sets. Since the test labels are not available, here we report results on the validation data.

\subsection{Settings}
As initial step, we performed some preprocessing on the images of both datasets. For Oulu-CASIA, we first detected the face using the method proposed in~\cite{viola2004robust}. For SFEW, we used the aligned faces provided by the dataset. Then, we detected $49$ facial landmarks on each face using the Chehra Face Tracker~\cite{asthana2014incremental}. All frames were cropped and resized to $224 \times 224$, which is the input size of the DCNN models.

\textbf{VGG fine-tuning:}
Since the two datasets are quite different, we fine-tuned the VGG-face model on each dataset separately. To keep the experiments consistent with~\cite{ding2017facenet2expnet} and~\cite{ofodile2017automatic}, we conducted ten-fold cross validation on Oulu-CASIA. This results in ten different deep models, each of them is trained on nine splits with $9 \times 3 \times(480 / 10) =1,296$ images. On the SFEW dataset, one model is trained using the provided training data. The training procedure for both datasets is executed for 100 epochs, with a mini-batch size of 64 and learning rate of $0.0001$ decreased by $0.1$ after $50$ epochs. The momentum is fixed to be $0.9$, and Stochastic Gradient Descent is adopted as optimization algorithm.
The fully connected layers of the VGG-face model are trained from scratch by initializing them with a Gaussian distribution. For data augmentation, we used horizontal flipping on the original data without any other supplementary datasets.

\textbf{ExpNet training:} Also in this case, a ten-fold cross validation is performed on Oulu-CASIA requiring the training of ten different deep models. The ExpNet architecture consists of five convolutional layers, each one followed by Relu activation and max pooling~\cite{ding2017facenet2expnet}. As mentioned in Section~\ref{sect:GDCNN}, these layers were trained first by regularization with the fine-tuned VGG model, then we appended one fully connected layer of size $128$. The whole network is finally trained. All parameters used in the ExpNet training (learning rate, momentum, mini-batch size, number of epochs) are the same as in~\cite{ding2017facenet2expnet}. 
We conducted all our training experiments using the Caffe deep learning framework~\cite{jia2014caffe}.

\textbf{Features extraction:}
We used the last pooling layer of DCNN models to extract features from each face image. This layer provides $512$ feature maps of size $7 \times 7$, which yields to covariance descriptors of size $512 \times 512$.
For the local approach, to well map landmarks position in the input image to the coordinates of the feature maps, we resized all feature maps to $14\times 14$, that allows us to correctly localize regions on the feature maps and minimize the overlapping between them.
The detected regions in the input image were mapped to the feature maps using Eq.~\eqref{eq:Mapping} with a ratio $s=1/16$. Based on this mapping, we extracted features around eyes, mouth and both cheeks from each feature map. Finally, we used these local features to compute a covariance descriptor of size $512 \times 512$ for each region in the input image. In Sections 1 and 2 of the supplementary material, we show images of the extracted global and local FMs and their corresponding covariance matrices. 

\textbf{Classification:}
For the global approach, each static image is represented by a covariance descriptor of size $512 \times 512$. In order to compare covariance descriptors in $Sym^{++}(512)$, it is empirically necessary to ensure their positive definiteness by using their regularized version, $C_{\Phi(f)} + \epsilon I$, where $\epsilon$ is a small regularization parameter (set to $0.0001$ in all our experiments), and $I$ is the $512 \times 512$ identity matrix.
To classify these descriptors, we used multi-class SVM with Gaussian kernel on the Riemannian manifold $Sym^{++}(512)$.
For reproducibility, we choose parameters of the Gaussian kernel $\gamma$ and SVM cost $\delta$ using cross validation with grid search in the following intervals: $\gamma \in [10^{-3}, 10^{-10}]$ and $\delta \in [10^{3}, 10^{8}]$.

Concerning the local approach, each image was represented by four covariance descriptors, each  regularized as stated for the global covariance descriptor. This resulted in four classification decisions that were combined using two late fusion methods: \textit{weighted sum} and \textit{product}. The best performance were achieved for weighted sum fusion with $w_{global}$, $w_{eyes}$, $w_{cheek-left}$, $w_{cheek-right}$ equal to $1$ and $w_{mouth}=0.2$, for the Oulu-CASIA dataset, and $w_{global}=1$, and $w_{eyes}$, $w_{mouth}$, $w_{cheek-left}$, $w_{cheek-right}$ equal to $0.1$ for the SFEW dataset. Note that we report the results of our local approach with only ExpNet model since it provides better results with the global approach than VGG-face model. SVM classification was obtained using the LIBSVM~\cite{chang2001libsvm} package.
Note that for testing the Oulu-CASIA dataset, we represented each video by its three peak frames as in Ding et al.~\cite{ding2017facenet2expnet}. Hence, to calculate the distance between two videos, we considered the mean of the distances between their frames. For softmax, we considered the video as correctly classified if its three frames are correctly recognized by the model.

\subsection{Results and discussion}
As first analysis, in Table~\ref{tab:Accuracy}, we compare our proposed global (G-FMs) and local (R-FMs) solutions with the baselines provided by the VGG-face and ExpNet models, without the use of the covariance matrix (\textit{i.e.}, they used the fully connected and softmax layers).
On Oulu-CASIA, the G-FMs solution improves by $3.7\%$ and $1.26\%$, respectively, the VGG-face and ExpNet models.
Though less marked, an increment of $0.69\%$ for the VGG-face and of $0.92\%$ for ExpNet has been also obtained on the SFEW dataset. These results prove that the covariance descriptors computed on the convolutional features provide more discriminative representations. Furthermore, the classification of these representations using Gaussian kernel on SPD manifold is more efficient than the standard classification with fully connected layers and softmax, even if these layers were trained in an end-to-end manner.
Table~\ref{tab:Accuracy} also shows that the fusion of the local (R-FMs) and global (G-FMs) approaches achieves a clear superiority on the Oulu-CASIA dataset surpassing by $1.25\%$ the global approach, while no improvement is observed on the SFEW dataset. This is due to the failure of landmark detection skewing the extraction of the local deep features. In Section~3 of the supplementary material, we show some failure cases of landmark detection on this dataset. 

\setlength{\tabcolsep}{4pt}
\begin{table}[!ht]
\begin{center}
\small
\begin{tabular}{lllll}
\hline
\textbf{Dataset} & \textbf{Model} & \textbf{FC-Softmax} & \textbf{ours (G-FMs)} & \textbf{ours (G-FMs and R-FMs)} \\

\hline

Oulu-CASIA & \it VGG Face & 77.8 & 81.5 & -- \\
& \it ExpNet & 82.29 & \textbf{83.55} & \textbf{84.80} \\

\hline

SFEW &\it VGG Face & 46.66 & 47.35 & -- \\
& \it ExpNet & 48.26 & \textbf{49.18} & \textbf{49.18} \\
\hline
\end{tabular}
\end{center}
\caption{Comparison of the proposed classification scheme with respect to the VGG-Face and ExpNet models with fully connected layer and Softmax.}
\label{tab:Accuracy}
\end{table}
\setlength{\tabcolsep}{1.4pt}

In Table~\ref{tab:ResultsRegions}, we investigated the performance of the individual regions of the face for ExpNet.
On both datasets, the right and left cheek provide almost the same score outperforming at a large extent the mouth score. Results for the eye region are not coherent across the two datasets: the eyes region is the best performing for Oulu-CASIA, but this is not the case on SFEW. We motivate this result by the fact that, in the wild acquisitions as for the SFEW dataset, the region of the eyes can be affected by occlusions, and the landmarks detection can be less accurate (see Section~3 of the supplementary material for failure cases of landmark detection in this dataset). 
Table~\ref{tab:ResultsRegions} also compares different fusion modalities. We found consistent results across the two datasets, indicating the weighted sum fusion between G-FMs and R-FMs is the best approach.

\setlength{\tabcolsep}{4pt}
\begin{table}[!ht]
\begin{center}
\small
\begin{tabular}{llll}
\hline
\textbf{Region}  & \textbf{Oulu-CASIA} & \textbf{SFEW}  \\

\hline

\it Eyes & 84.59 & 38.05
\\
\it Mouth & 70.00 & 38.98 \\

\it Right Cheek&  83.96 & 43.16 \\
\it Left Cheek &  83.12 & 42.93 \\

\hline

\it R-FMs product fusion &  83.66 & 42.92 \\
\it G-FMs and R-FMs product fusion & 84.05& 45.24 \\

\hline

\it R-FMs weighted-sum fusion & 84.59 & 43.85 \\
\it \textbf{G-FMs and R-FMs weighted-sum fusion} & \textbf{84.80} & \textbf{49.18} \\

\hline

\end{tabular}
\end{center}
\caption{Overall accuracy (\%) of different regions and fusion schemes on the Oulu-CASIA and SFEW datasets for the ExpNet model.}
\label{tab:ResultsRegions}
\end{table}
\setlength{\tabcolsep}{1.4pt}

\begin{figure}[!ht]
\centering
\begin{minipage}{0.4\linewidth}
\centering
\includegraphics[width=\linewidth]{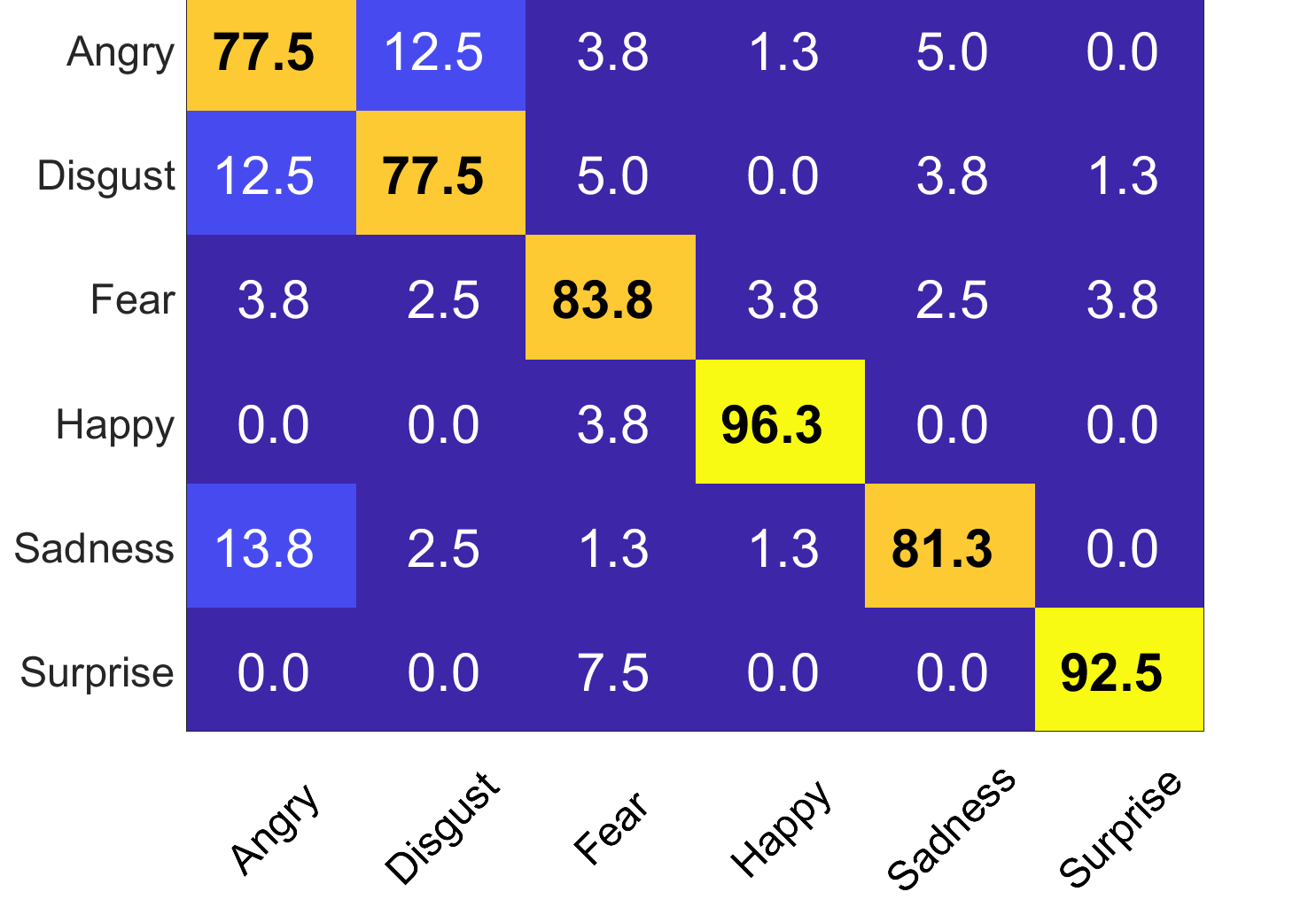}
\end{minipage}
\begin{minipage}{0.4\linewidth}
\centering
\includegraphics[width=\linewidth]{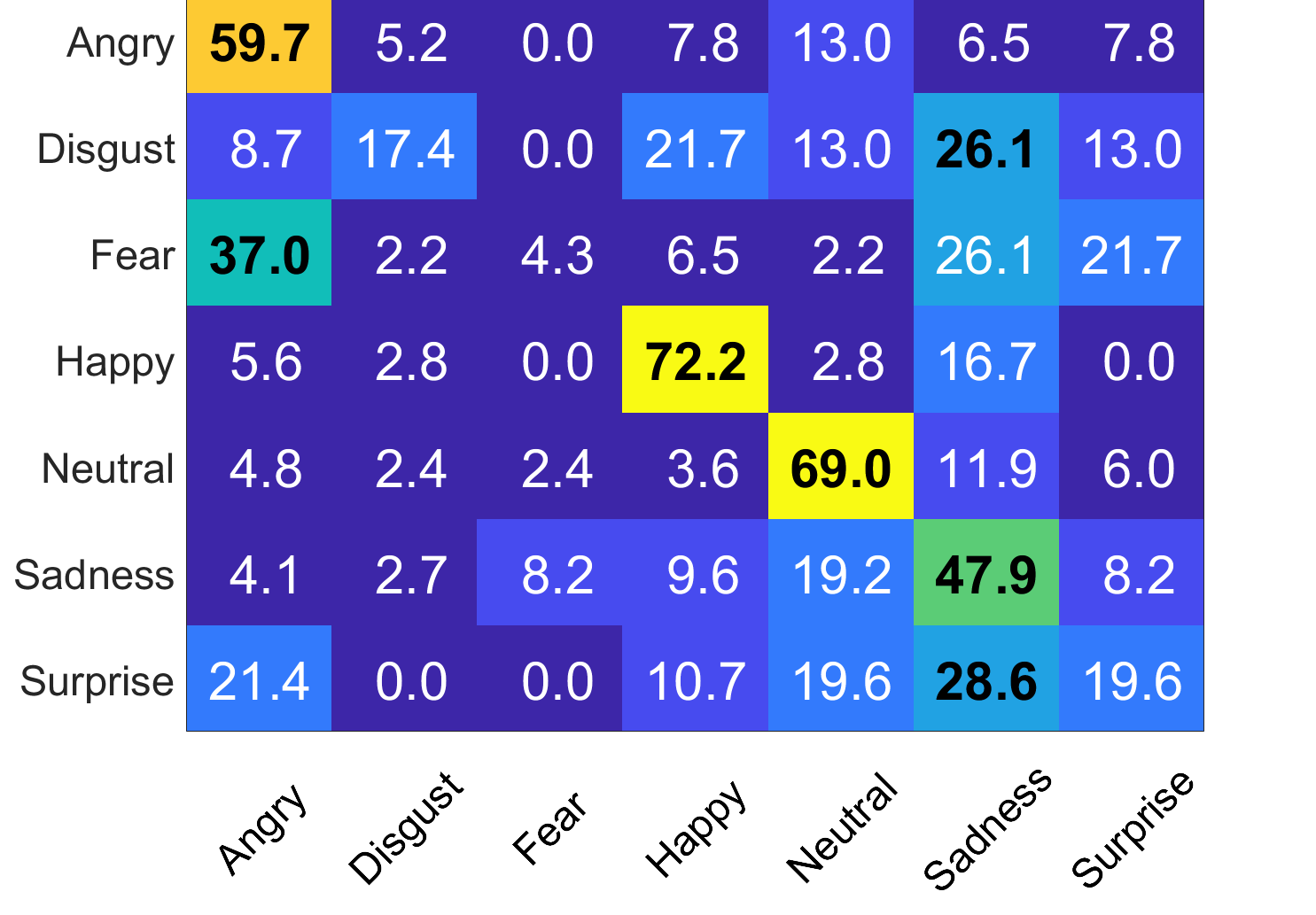}
\end{minipage}
\caption{Confusion matrix on Oulu-CASIA (left) and SFEW (right) for ExpNet with weighted-sum fusion.}
\label{fig:sfew_conf}
\end{figure}

The confusion matrices for ExpNet with weighted-sum are reported in Figure~\ref{fig:sfew_conf} left and right plots, respectively, for Oulu-CASIA and SFEW. For Oulu-CASIA, the happy and surprise expressions are better recognized over the rest. The happy expression is the best recognized one also for SFEW, followed by the neutral one. 

\setlength{\tabcolsep}{4pt}
\begin{table}[!ht]
\begin{center}
\small
\begin{tabular}{llll}
\hline
\textbf{Method} & \textbf{Oulu-CASIA} & \textbf{SFEW} \\
\hline
\it Kacem et al.~\cite{kacem:2017}~$^*$ & 83.13 & --  \\
\it Jung et al.~\cite{jung2015joint}~$^*$ & 74.17 & -- \\
\hline
\it Liu et al.~\cite{liu2013aware} & -- & 26.14 \\
\it Levi et al.~\cite{levi2015emotion}& -- & 41.92 \\
\it Mollahosseini et al.~\cite{mollahosseini2016going} & -- & 47.70 \\
\it Ng et al.~\cite{ng2015deep} & -- & 48.50\\
\it Yu et al.~\cite{yu2015image} & --& 52.29\\
\it Ding et al.~\cite{ding2017facenet2expnet} & 82.29 & 48.29 \\
\it  Liu et al.~\cite{liu2014learning}~$^*$ & 74.59 & -- \\
\it Guo et al.\cite{guo2012dynamic}~$^*$ & 75.52 & -- \\
\it Zhao et al.~\cite{zhao2016peak}~$^*$  & 84.59 & -- \\
\hline
\it Jung et al.~\cite{jung2015joint}~$^*$ & 81.46 & -- \\
\it Ofodil et al.~\cite{ofodile2017automatic}~$^*$ & 89.60 & -- \\

\hline
\it \textbf{ours (ExpNet + G-FMs)} & \textbf{83.55} & \textbf{49.18} \\
\it \textbf{ours (ExpNet + G-FMs and R-FMs fusion)} & \textbf{84.80} &
\textbf{49.18} \\
\hline
\end{tabular}
\end{center}
\caption{Comparison with state-of-the art solutions on Oulu-CASIA and SFEW. Geometric, appearance and hybrid solutions are reported in the first three groups of methods; Our solutions are given in the last row. ($^*$) Dynamic approaches.}
\label{tab:ResultsComparisons}
\end{table}

As last analysis, in Table~\ref{tab:ResultsComparisons} we compare our solution with respect to state-of-the-art methods.
Overall, on Oulu-CASIA, we obtained the second highest accuracy, outperforming several recent approaches. Furthermore, Ofodil et al.~\cite{ofodile2017automatic}, who achieved the highest accuracy on this dataset, also used temporal information of the video. In addition, they did not report the frames used to train their DCNN model, which is indeed an important information to compare the two approaches. Note that, to compare our results with those of Ding et al.~\cite{ding2017facenet2expnet}, which was reported per frames, we reproduced the results for their approach on a per video basis, considering that the video is correctly classified if the three frames of the video are correctly recognized.
On the SFEW dataset, the global approach achieves the second highest accuracy, surpassing various state of the art methods with significant gains. Moreover, the highest accuracy reported by~\cite{yu2015image} is obtained using a DCNN model trained on more than $35,000$ additional data provided by the FER-2013 database~\cite{goodfellow2013challenges}. As reported in~\cite{ding2017facenet2expnet}, this data augmentation can improve results on SFEW from $48.29\%$ to $55.15\%$. 
\section{Conclusions}\label{sect:conclusion}
In this paper, we have proposed the covariance matrix descriptor as a way to encode DCNN features in facial expression recognition. 
The covariance matrix belongs to the set of symmetric positive-definite (SPD) matrices, thus laying on a special Riemannian manifold. We have shown the classification of these representations using Gaussian kernel on the SPD manifold is more efficient than the standard classification with fully connected layers and softmax. By implementing our approach using different architectures, \emph{i.e.}, VGG-face and ExpNet, in extensive experiments on the Oulu-CASIA and SFEW datasets, we have shown that the proposed approach achieves state-of-the-art performance for facial expression recognition. As future work, we aim to include the temporal dynamics of the face in the proposed model.



\newpage








\begin{center}
{\Large \textbf{A Supplementary Material to the Paper: Deep Covariance Descriptors for Facial Expression Recognition\\}}
\end{center}
In this supplementary material, we present further details on the conducted experiments. 
In particular, we provide visualizations of: \\

\begin{figure}[!ht]
\centering
\includegraphics[width=0.9\linewidth]{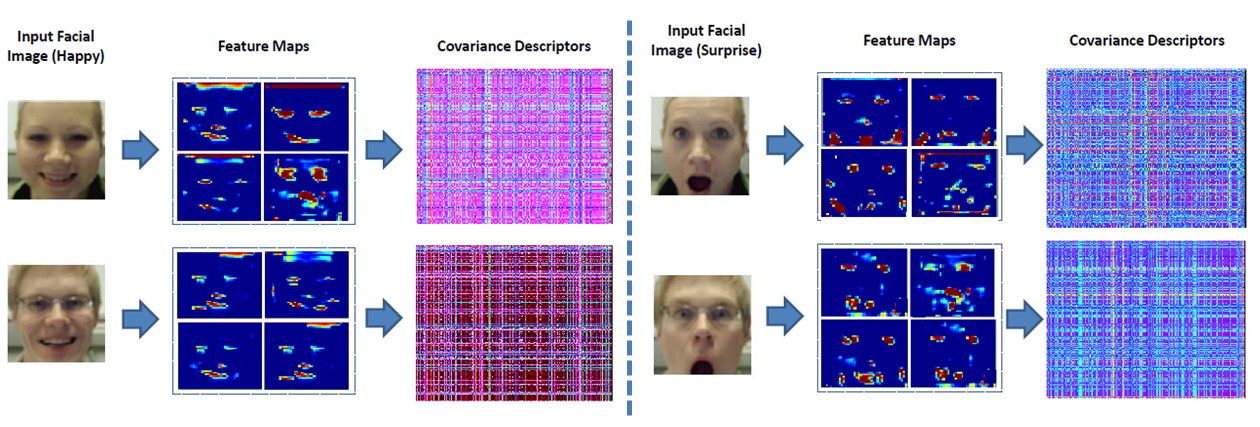}
\caption{Visualization of some feature maps (ExpNet) and their corresponding covariance descriptors for two subjects from the Oulu-CASIA dataset conveying happy and surprise expressions. We show four feature maps (chosen from 512 feature maps) for each example image. The corresponding covariance descriptors are computed over the 512 FMs. Best seen in color.}
\label{fig:feature-maps}
\end{figure}

 \textbf{Global DCNN features and their covariance descriptors:} 
 Figure~\ref{fig:feature-maps} shows four selected feature maps (chosen from 512 FMs) extracted with the ExpNet model for two subjects of the Oulu-CASIA dataset (happy and surprise expressions). We also show the global covariance descriptor relative to the 512 feature maps as a 2D image. Common patterns can be observed in the covariance descriptors computed for similar expressions, \emph{e.g.}, the dominant colors in the covariance descriptors of happy expression (left panel) are pink/purple, while being blue in the covariance descriptors of surprise expression (right panel). 

\begin{figure}[!ht]
\centering
\includegraphics[width=0.8\linewidth]{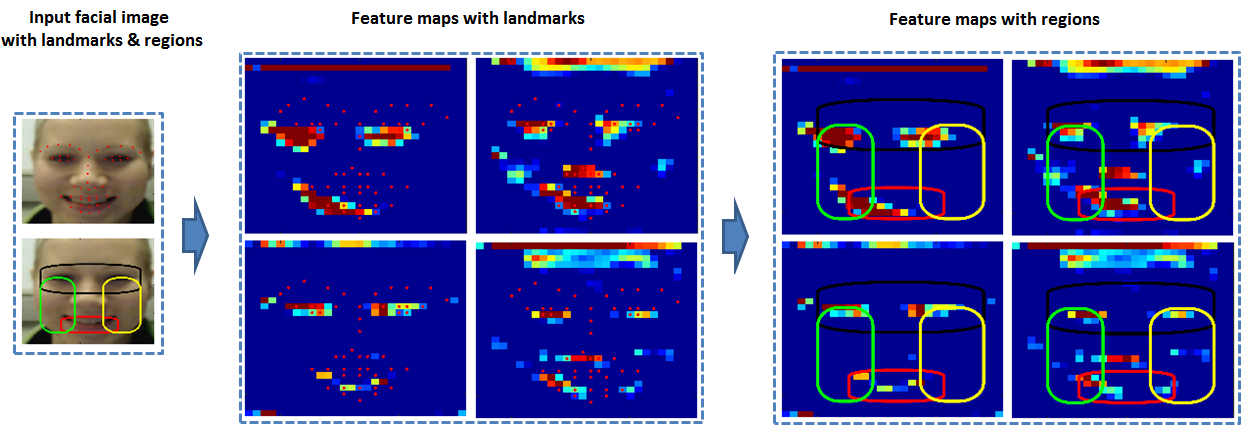}
\caption{Visualization of the detected facial landmarks and regions on the input facial image and mapped on four selected feature maps (from 512). Best seen in color.}
\label{fig:RegionFMs}
\end{figure}

\begin{figure}[!ht]
\centering
\includegraphics[width=0.9\linewidth]{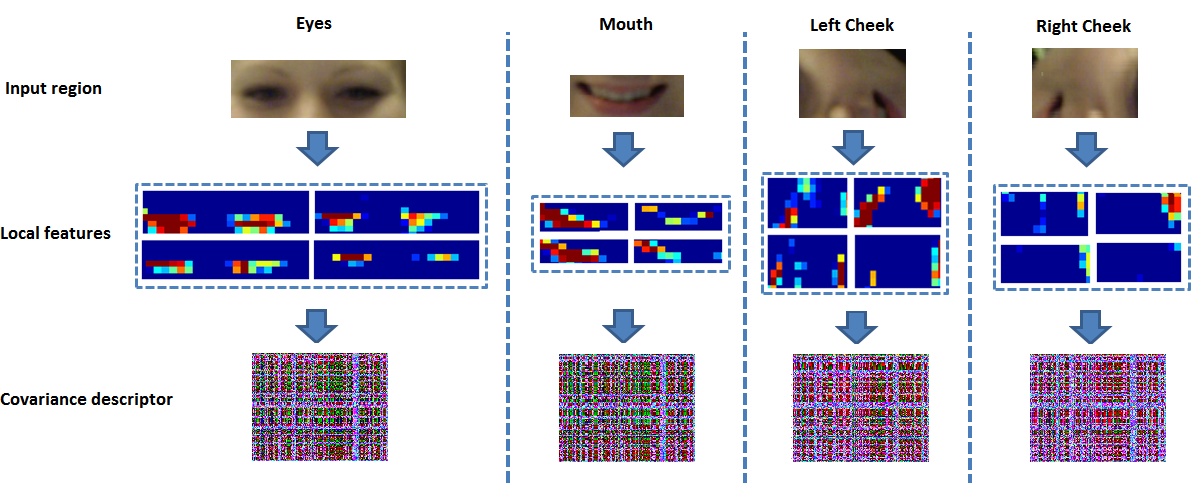}
\caption{Visualization of the extracted regions on four feature maps and their corresponding covariance descriptors. Best seen in color.}
\label{fig:RegionCovs}
\end{figure}

\textbf{Local DCNN features and their covariance descriptors:} Figure~\ref{fig:RegionFMs} shows the four local regions detected on the input facial image on the left; then, landmarks and regions are shown on four selected feature maps, as mentioned in Section~3.2 of the paper. These FMs are selected from 512. 
The covariance descriptors relative to each detected region are shown in Figure~\ref{fig:RegionCovs}. We can observe that each local covariance descriptor captures different patterns. 

\begin{figure}[!ht]
\centering
\includegraphics[width=.8\linewidth]{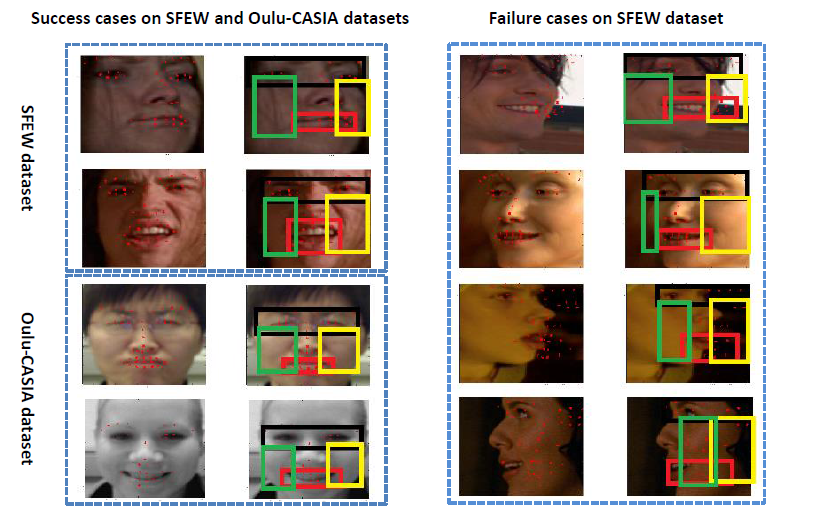}
\caption{Examples of facial landmark and region detection on the SFEW and Oulu-CASIA datasets, with some failure cases for the SFEW dataset. For each example, the image on the left shows the aligned face with its landmark points, while the image on the right represents the aligned face with its detected regions.}
\label{fig:landfail}
\end{figure}

\textbf{Failure cases of facial landmark detection on SFEW dataset:} 
Figure~\ref{fig:landfail} exhibits some failure and success cases of facial landmark and region detection on the input facial images. In the left panel of this figure, we show examples from the Oulu-CASIA and SFEW datasets, where the landmark and region detection succeeded. In the right panel, we show four failure examples for landmark and region detection in the SFEW dataset. We noticed that this step failed on $\sim30\%$ of the facial images of SFEW. This explains why we do not obtain improvements by combining local and global covariance descriptors on this dataset. 

\end{document}